\documentclass[journal, onecolumn, 11pt]{IEEEtran}
\IEEEoverridecommandlockouts

\usepackage{makecell}
\usepackage{xcolor}
\def\BibTeX{{\rm B\kern-.05em{\sc i\kern-.025em b}\kern-.08em
    T\kern-.1667em\lower.7ex\hbox{E}\kern-.125emX}}
    
\usepackage{amsmath,amssymb,framed}

\usepackage{epsfig,graphics}
\usepackage{multicol,subfigure}
\usepackage{wrapfig, rotating,setspace}
\usepackage[color=yellow]{pdfcomment}
\usepackage{colortbl,hhline}
\usepackage{wrapfig}
\usepackage{setspace} 
\usepackage{multirow}
\usepackage{algorithm,algorithmic}
\usepackage{booktabs}
\usepackage{comment}
\usepackage[mathscr]{eucal}
\usepackage{amsbsy}
\usepackage{bm}
\MakeRobust{\overrightarrow}

\usepackage{comment}

\usepackage[mathscr]{eucal} 
\usepackage{cite}
\usepackage{amsbsy}
\graphicspath{ {./figure/} } 
\usepackage{bm}

\newtheorem{theorem}{Theorem}

\newtheorem{definition}{Definition}

\begin{document}

\title{Soft and subspace robust multivariate rank tests based on entropy regularized optimal transport\\
\thanks{Shuchin Aeron is supported by NSF CCF:1553075, NSF TRIPODS grant HDR:1934553, and AFOSR FA9550-18-1-0465., and US Army Natick Center \& Tufts CABCS. Shoaib Bin-Masud is exclusively supported by US Army Natick Center \& Tufts CABCS. Boyang Lyu is exclusively supported by AFOSR FA9550-18-1-0465. Correspondence to: shoaib.masud@tufts.edu, shuchin@ece.tufts.edu}
}

\author{\IEEEauthorblockN{Shoaib Bin Masud, Boyang Lyu, Shuchin Aeron}\\
\IEEEauthorblockA{\textit{Dept. of ECE, \textit{Tufts University}}}
}

\maketitle

\begin{abstract}
In this paper, we extend the recently proposed multivariate rank energy distance, based on the theory of optimal transport, for statistical testing of distributional similarity, to soft rank energy  distance. Being differentiable, this in turn allows us to extend the rank energy to a subspace robust rank energy distance, dubbed Projected soft-Rank Energy  distance, which can be computed via optimization over the Stiefel manifold. We show via experiments that using projected soft rank energy one can trade-off the detection power vs the false alarm via projections onto an appropriately selected low dimensional subspace. We also show the utility of the proposed tests on unsupervised change point detection in multivariate time series data. All codes are publicly available at the link provided in the experiment section. 
\end{abstract}

\section{Introduction}
This paper is concerned with the classical two-sample hypothesis test, also known as the Goodness-of-Fit (GoF) test. 
Given samples $\bm{X}_1, \dots, \bm{X}_m$ and $\bm{Y}_1, \dots, \bm{Y}_n$, that are independent and identically distributed (i.i.d.) according to distributions $\mu$ and $\nu$, respectively, on $\mathbb R^d$, one wants to test following hypotheses, 
$$H_0: \mu = \nu, \hspace{2mm}\text{vs.}\hspace{2mm} H_1: \mu \neq \nu. $$
Two sample multivariate goodness-of-fit-test has been studied extensively using both parametric and non-parametric methods. A brief and relevant background is provided in Section \ref{sec:background}, but we refer the reader to some classical \cite{bickel1969distribution, weiss1960two, friedman1979multivariate, schilling1986multivariate, wasserman2013all} and recent texts \cite{baringhaus2004new, sejdinovic2013equivalence, szekely2013energy, wasserman2006all} for background. To be concise and retain focus on the approach undertaken in this paper, we begin by summarizing the main contributions below. To fully appreciate the contributions, the reader may directly peruse section \ref{sec:background} first and come back here.

\subsection{Summary of main contributions}
\begin{enumerate}
    \item We extend the recently proposed nonparametric two-sample testing proposed in \cite{deb2019multivariate} to a new test that exploits the recent developments in computation of optimal transport, namely entropic relaxation \cite{peyre2019computational}. We show via extensive experiments that this proposed test is nearly non-parametric under the null for small values of entropic regularization parameter. 
    \item In contrast to \cite{deb2019multivariate}, the proposed test provides an efficient and scalable approximation to optimal transport, always admissible, a differential function of the data and we exploit this fact towards proposing a subspace robust two-sample testing that has the benefit to trade-off power of detection and false alarms in high dimensional testing under a limited number of samples. In this context, we provide an approach that is complementary to subspace robust Wasserstein distances and tests based on that \cite{paty2019subspace, wang2020two}. This test can be efficiently computed using a Riemannian optimization over the Stiefel manifold. 
    \item We provide an application of the proposed test to unsupervised detection of change points in multivariate time series data and observe that the test is very sensitive to shifts in support of the signal. This is a useful property but it makes the test also very sensitive to the presence of outliers.  
\end{enumerate}

The rest of the paper is organized as follows. In section \ref{sec:background} we provide the context needed to appreciate the recent developments and the subsequent innovations made in this paper. In section \ref{sec:proposed} we outline in detail the proposed tests. In section \ref{sec:sims} we provide ample empirical evidence in support of the proposed tests and an application to unsupervised detection of change points in multivariate time series data.

\section{Background and related work}
\label{sec:background}
\textbf{Notation}: Let $\mathcal{P}(\mathbb{R}^d)$ denote the set of all probability measures on $\mathbb{R}^d$. Unless otherwise stated we will assume that distributions $\mu, \nu,...$ are absolutely continuous. For a random variable $\bm{X}$, $\bm{X} \sim \mu$ reads that $\bm{X}$ is distributed with the distribution $\mu$. We will use upper-case letters $X,Y$ for univariate random-variables. The notation $\mathbf{T, P}$ will be reserved for maps, matrices, etc. The rest of the notation is introduced as needed.

\subsection{Multivariate Ranks and Quantiles} 

Rank based distribution-free goodness of tests have been studied comprehensively in 1-D e.g Kolmogorov-Smirnov test\cite{smirnov1939estimation},  Wilcoxon  signed-rank test\cite{wilcoxon1947probability},  Wald-Wolfowitz  runs  test \cite{wald1940test}. Lack of canonical ordering in $d$- dimensional space, for $d\geq 2$, does not guarantee the exact distribution freeness studied in \cite{bickel1965some, puri1966class, chaudhuri1996geometric, ghosh1999multivariate, liu1993quality, hallin2004rank, hallin2006parametric}. Recently  a number of authors \cite{hallin2017distribution,hallin2021distribution, chernozhukov2017monge,deb2019multivariate,hallin2020fully, shi2020rate, shi2020distribution} studied the notion of multivariate rank based on OT theory. In this paper we will consider the setting in \cite{deb2019multivariate} explicitly and build upon the ideas therein.

\paragraph{Ranks and Quantiles for univariate distributions} Let $X$ be a univariate random variable with c.d.f. $\mathsf F: \mathbb{R} \rightarrow [0,1]$. It is a standard result that when $\mathsf F$ is continuous, the random variable $\mathsf{F}(X) \sim \mathsf{U} [0, 1]$ - the uniform distribution on $[0,1]$. For any $x \in \mathbb{R}$, $\mathsf{F}(x)$ is referred to as the \textit{rank-function}. For any $0<p<1$, the \emph{quantile function} is defined by $\mathsf{Q}(p) = \inf \{x \in \mathbb R: p \leq \mathsf F(x)\}$. When $\mathsf F$ is continuous, the quantile function $\mathsf Q = \mathsf F^{-1}$. 

\paragraph{Ranks and Quantiles for multivariate distributions}  Since there exists no natural ordering in $\mathbb R^d$, defining ranks and quantiles in high dimension is not straightforward. To extend the notion of rank in $\mathbb R^d$, theory of Optimal Transport (OT) has been used to propose meaningful and useful notions of multivariate rank and quantile functions \cite{hallin2017distribution,chernozhukov2017monge,deb2019multivariate,hallin2020fully}. In it's most standard setting, given two distributions, a source distribution $\mu\in \mathcal P(\mathbb R^d)$ and a target distribution $\nu \in \mathcal P(\mathbb R^d)$, OT aims to find a map $ \mathbf{T}: \mathbb R^d \rightarrow \mathbb R^d$ that pushes $\mu$ to $\nu$  with a minimal cost. That is, given $\bm X\sim \mu$, and $\bm{Y} \sim \nu $, OT finds a map $\mathbf{T}$ that solves for, 
\begin{equation}\label{monge}
    \inf_{\mathbf T} \int \|\bm{x} - \mathbf T(\bm{x})\|^2 d\mu(\bm{x}) \;\; \text{subject to.}\;\; \bm{Y} = \mathbf{T}(\bm{X}) \sim \nu  
\end{equation}
Note that if $\mathbf T(\bm{X}) \sim \nu$ when $\bm{X} \sim \mu$, we write $\nu = \mathbf{T}_{\#} \mu$. The \textit{key insight} in using the theory of OT to multivariate ranks and quantiles, comes from noticing that in case of $d = 1$, the optimal transport map is given by $\mathbf T = \mathsf F_\nu^{-1} \circ \mathsf F_\mu$ where $\mathsf F_\mu$ and $\mathsf F_\nu$ are the distribution functions for $\mu$ and $\nu$ respectively. When $\nu = \mathsf U[0, 1]$, this gives the rank function $\mathsf F_\mu$.  Following  \cite{deb2019multivariate}, McCann's theorem \cite{mccann1995existence} stated below, is used to extend the notion of rank to the multivariate setting.

\begin{theorem}[McCann~\cite{mccann1995existence}] \label{Theorem 1}
Assume $\mu, \nu \in \mathcal P(\mathbb R^d)$ be absolutely continuous measures, then there exists transport maps $\mathbf{R}(\cdot)$ and $\mathbf{Q}(\cdot)$, that are gradients of real-valued $d$-variate convex functions such that $\mathbf{R}_\# \mu =\nu, \;\; \mathbf{Q}_\#\nu = \mu$, $\mathbf{R}$ and $\mathbf{Q}$ are unique and $\mathbf{R}\circ \mathbf{Q}(\bm{x}) = \bm{x}$, $\mathbf{Q}\circ \mathbf{R}(\bm{y}) = \bm{y}$.
\end{theorem}
Based on this result, the authors in \cite{deb2019multivariate} give the following definitions for the rank and quantile functions in high dimensions. 
\begin{definition}[Deb~ \cite{deb2019multivariate}]
Given absolutely continuous measure $\mu \in \mathcal P(\mathbb R^d)$ and $\nu =\mathsf U[0,1]^d$ - the uniform measure on the unit cube in $\mathbb{R}^d$, the ranks and quantile \textit{maps} for $\mu$ are defined as the maps $\mathbf{R}(\cdot)$ and $\mathbf{Q}(\cdot)$ respectively as defined in Theorem \ref{Theorem 1}.
\end{definition}

\subsection{Multivariate Rank from Samples}
Let $\bm{X}_1, \cdots, \bm X_n$ be \emph{i.i.d.} samples drawn from $\mu \in \mathcal P(\mathbb R^d)$. Let $\mu_n$ denote the empirical measure supported on these samples. In general, there does not exist a map $\mathbf{R}$ from $\mu_n$ to $\mathsf U[0,1]^d$\footnote{While there does not exist a map, there does exist a \textit{plan} obtained by solving the so-called Kantorovich problem, which is a relaxation of the Monge problem \cite{chernozhukov2017monge}.} On the other hand there does exist a map $\mathbf{Q}$ from $\mathsf U [0,1]^d$ to $\mu_n$, and is a special case of the OT problem referred to as the semi-discrete optimal transport. Using the convex geometry of the problem the authors in \cite{ghosal2019multivariate} propose a  \textit{randomized} rank map by using the solution to the Quantile map $\mathbf{Q}$ obtained from the semi-discrete OT. However, it may be computationally expensive to do so \cite{ghosal2019multivariate}. To circumvent this, the authors in \cite{deb2019multivariate} propose to use the following notion of empirical multivariate rank, by mapping the samples $\bm{X}_1, \cdots, \bm X_n$ to fixed set $d$-dimensional Halton sequence \cite{chi2005optimal} of size $n$, denoted by $\mathcal H_n^d:=\{\bm{h}_1,\dots, \bm{h}_n\}.$ For details on Halton sequences we refer the reader to \cite{chi2005optimal}. The main point to note is that the empirical measure $\nu_n = \frac{1}{n}\sum_{ i = 1}^n \delta_{\bm h_i}$ converges in distribution to $\mathsf U[0,1]^d$.

Now given the  empirical distribution of samples and the empirical distribution concentrated on a given set of Halton sequences, 
$$\mu_n^{\bm{X}} = \frac{1}{n}\sum_{ i = 1}^n \delta_{\bm{X}_i}\;\; \text{and}\;\; \nu_n^{\bm{H}} = \frac{1}{n}\sum_{ i = 1}^n \delta_{\bm h_i},$$ the empirical rank function is defined as follows. First, one solves for the following discrete optimal transport problem, also known as the 2-D assignment problem in the literature \cite{peyre2019computational}. 

\begin{align}
\label{eq:OT}
    \hat{\mathbf{P}}_n = \arg \min_{\mathbf{P} \in \Pi} \sum_{i,j = 1}^{n} \mathbf{C}_{i,j} \mathbf{P}_{i,j},
\end{align}
where $\mathbf{C}_{i,j} = \| \bm{X}_i - \bm{h}_j\|^2$, $\Pi = \{ \mathbf{P}: \mathbf{P} \bm{1} = \frac{1}{n} \bm{1}, \bm{1}^\top \mathbf{P} = \frac{1}{n} \bm{1}^\top\}$. It is well known that the solution to this problem, under the given set-up, is one of the scaled permutation matrices. Consequently, one obtains a map $\hat{\mathbf{R}}_n(\bm{X}_i) = \bm{h}_{\sigma(i)}$, where $\sigma(i)$ is the non-zero index in the $i$-th row of $\hat{\mathbf{P}}_n$.
\begin{definition}\cite{deb2019multivariate}  $\hat{\mathbf{R}}_n$ as defined above is called as the empirical rank map.
\end{definition}

\subsection{Goodness-of-Fit (GoF) tests: Rank Energy Distance)}
Given two sets of i.i.d. samples $\{\bm X_1, \dots, \bm X_m\}\in \mathbb R^d$ and $\{\bm Y_1, \dots, \bm Y_m\}\in \mathbb R^d$ with empirical measures $\mu_m^{\bm X}$, $\mu_n^{\bm Y}$ respectively. In \cite{deb2019multivariate}, the authors proposed a distribution-free multivariate GoF test based on the energy distance \cite{szekely2013energy} between the empirical ranks obtained for each of the sample sets, which is referred as the Rank Energy (RE) test. Following \cite{deb2019multivariate}, $\mathsf{RE}$ for GoF testing is defined as follows. Draw $m+n$ Halton sequences $\bm{h}_1,\cdots, \bm{h}_{m+n}$ and compute the joint-empirical rank map $\hat{\mathbf{R}}_{m,n}$ between the empirical measure formed by combining the two sets of samples, 
$$\mu_{m,n}^{\bm{X}, \bm{Y}} = \frac{1}{m + n} ( m \mu_{n}^{\bm{X}} + n \mu_{m}^{\bm{Y}}),$$

and the empirical measure concentrated on the Halton sequences, $ \nu_{m,n}^{\mathbf{H}} = \frac{1}{m + n}\sum_{ i = 1}^{ m + n} \delta_{\bm h_i}$. The rank energy is defined as, 
\begin{align}\label{re}
    \mathsf{RE}_{m,n}^2 \doteq & \frac{2}{mn} \sum_{i=1}^{m} \sum_{j=1}^{n} \| \hat{\mathbf{R}}_{m,n}(\bm{X}_i) - \hat{\mathbf{R}}_{m,n}(\mathbf{Y}_j) \| - \frac{1}{m^2} \sum_{i,j=1}^{m} \| \hat{\mathbf{R}}_{m,n}(\bm{X}_i) - \hat{\mathbf{R}}_{m,n}(\bm{X}_j)\| \notag \\ & - \frac{1}{n^2}  \sum_{i,j=1}^{n} \| \hat{\mathbf{R}}_{m,n}(\bm{Y}_i) - \hat{\mathbf{R}}_{m,n}(\bm{Y}_j)\|.
\end{align}
It can be shown that the $\mathsf{RE}_{m,n}^2$ is distribution free for all sample sizes, see \cite{deb2019multivariate}, which follows from the fact that under the null, the empirical ranks map is uniformly distributed over the $(m + n)!$ permutations. 

Based on \cite{deb2019multivariate}, the rank energy test  rejects   $H_0$ if $\frac{mn}{m + n} \mathsf{RE}_{m,n}^2 > \kappa_\alpha^{m,n}$, where $\kappa_\alpha^{m,n}$ is a threshold that is picked as a function of $m,n$ and a desired level of False Alarm (FA) level $\alpha$.

In general, in \eqref{re}, one can use a characteristic kernel $k(\cdot,\cdot)$ \cite{phillips2011gentle} and via defining
\begin{gather*}
    B = \frac{1}{mn}\sum_{i, j = 1}^{m, n} k(\hat{\mathbf{R}}_{m,n}(\bm{X}_i), \hat{\mathbf{R}}_{m,n}(\bm{Y}_j)),\;\;    C = \frac{1}{m^2}\sum_{i, j = 1}^{m} k(\hat{\mathbf{R}}_{m,n}(\bm{X}_i), \hat{\mathbf{R}}_{m,n}(\bm{X}_j)),\\
    D = \frac{1}{n^2}\sum_{i, j = 1}^{n} k(\hat{\mathbf{R}}_{m,n}(\bm{Y}_i),          \hat{\mathbf{R}}_{m,n}(\bm{Y}_j)), 
\end{gather*}
one can define a kernelized rank energy via the following equation:
\begin{align}
    \label{kre}
    \mathsf{RE}_{m,n}^2 = 2B - C - D.
\end{align}

\section{Proposed multivariate ranks and tests of goodness-of-fit}
\label{sec:proposed}

In this work we are motivated by  \textit{potential use} of these tests towards generative adversarial networks (GANs), which necessitates that the $\mathsf {RE}_{m,n}$ be a differentiable function of data samples $\bm{X}_1, \cdots, \bm{X}_m$, $\bm{Y}_1, \cdots, \bm {Y}_n$. In this context we are motivated by the use of Sinkhorn iterations for solving the discrete OT problem \cite{cuturi2013sinkhorn} via \textit{entropic regularization} and noting that as an algorithm one can readily differentiate through the iterative steps \cite{feydy2019interpolating}.

\subsection{Soft Rank Energy Distance}\label{sRE}
To define soft rank energy, we first need to define the notion of a soft rank.  To do so we consider the Optimal Transport \emph{Plan} between $\mu_{m}^{\bm{X}}$ and $\nu_{m}^{\bm{H}}$, via solving for the following entropic regularized problem \cite{peyre2019computational}, 
\begin{align}
\label{eq:OT_reg}
    \mathbf{P}^0 = \arg \min_{\mathbf{P} \in \Pi} \sum_{i,j = 1}^{m} \mathbf{C}_{i,j} \mathbf{P}_{i,j} - \epsilon H(\mathbf{P}),
\end{align}
where $\mathbf{C}_{i,j} = \| \bm{X}_i - \bm{h}_j\|^2$, $\epsilon > 0$,  $\Pi = \{ \mathbf{P}: \mathbf{P} \bm{1} = \frac{1}{n} \bm{1}, \bm{1}^\top \mathbf{P} = \frac{1}{n} \bm{1}^\top\}$, and $H(\mathbf{P}) = - \sum_{i,j} \mathbf{P}_{i,j} \log \mathbf{P}_{i,j}$ is the entropy functional. For $\epsilon >0$, $\mathbf{\hat R}_n$ will not be a permutation matrix, see \cite{peyre2019computational}, and in this case we define a soft rank map via, 
\begin{align}\label{softrank}
    \hat{\mathbf{R}}^{s,\epsilon} (\bm{X}_i) = \sum_{j = 1}^{m} \frac{\mathbf{P}_{i,j}^0}{\sum_{j=1}^{m} \mathbf{P}_{i,j}^0} \bm{h}_j.
\end{align}
Using this notion of rank, we define the soft rank energy statistics as follows. 
\begin{align}\label{sre}
    \mathsf{sRE}_{m,n}^2 \doteq & \frac{2}{mn} \sum_{i=1}^{m} \sum_{j=1}^{n} \| \hat{\mathbf{R}}_{m,n}^{s,\epsilon}(\bm{X}_i) - \hat{\mathbf{R}}_{m,n}^{s,\epsilon}(\bm{Y}_j) \| - \frac{1}{m^2} \sum_{i,j=1}^{m} \| \hat{\mathbf{R}}_{m,n}^{s,\epsilon}(\bm{X}_i) - \hat{\mathbf{R}}_{m,n}^{s,\epsilon}(\bm{X}_j)\| \notag \\ & - \frac{1}{n^2}  \sum_{i,j=1}^{n} \| \hat{\mathbf{R}}_{m,n}^{s,\epsilon}(\bm{Y}_i) - \hat{\mathbf{R}}_{m,n}^{s, \epsilon} (\bm{Y}_j)\|,
\end{align}
where $\hat{\mathbf{R}}_{m,n}^{s,\epsilon}$ is the map obtained by the entropic regularized mapping between $\mu_{m,n}^{\mathbf{X}, \mathbf{Y}}$ and $\nu_{m,n}^{\mathbf{H}}$. One can also use the kernelized version of $\mathsf{sRE}^2_{m,n}$ similar to $\mathsf{RE}^2_{m,n}$ (see equation \eqref{kre}).

\subsection{Projected Rank Energy Tests}\label{proj_sec}
Recently projected Wasserstein distances or subspace-robust Wasserstein distances were proposed (almost at the same time in different contexts) in \cite{paty2019subspace,niles2019estimation, deshpande2019max} for robustness to noise in computing the transport maps. This approach was recently exploited in \cite{wang2020two} for two-sample testing to address the diminishing power of the OT based tests in high-dimensions by finding an appropriate low-dimensional subspace to compare the two samples. In the present context, given two empirical distributions, $\mu_{m}^{\mathbf{X}}, \nu_{m}^{\mathbf{Y}}$, one may consider the following projected Wasserstein-2 distance is defined as, 
\begin{align}
    \label{PW}
    \max_{\mathbf{U} \in \mathbb{R}^{d \times k}: \mathbf{U}^\top \mathbf{U} = \mathbf{I}_k} \min_{\mathbf{P} \in \Pi(\mathbf{U}_{\#}\mu_m^{\bm X}, \mathbf{U}_{\#}\mu_n^{\bm Y})} \left(\sum_{i,j} \mathbf{P}_{i,j}\| \mathbf{U} \bm{X}_i - \mathbf{U} \bm{Y}_j\|^2 \right)
\end{align} 
Heuristically, we will assume that with high probability the projection will not change the weights of the empirical distribution. Therefore, the marginal constraint on the coupling in the inner minimum does not change with $\mathbf{U}$. Motivated by these developments, here we propose the subspace robust versions of soft rank energy, namely the projected rank energy statistics.

We first define Projected soft Rank Energy Distance ($\mathsf{PsRE}^2_{m,n}$) as follows. Let $\hat{\mathbf{R}}_{m,n}$ be the empirical rank computed with projected data $\mathbf{U} \mathbf{U}^\top \bm{X}_i, \mathbf{U} \mathbf{U}^\top \bm{Y}_j, i= 1,\cdots, m, j = 1, \cdots, n$ and fixed set of Halton sequences $\bm{h}_1, \cdots \bm{h}_{m+n} \subset [0, 1]^{d}$ - in the unit cube in $\mathbf{R}^d$. In order that the Rank function still makes sense, i.e. McCann's theorem is still applicable, we add a small amount of Gaussian noise to the projected samples, $\mathbf{U} \mathbf{U}^\top \bm{X}_i + \sigma \bm{N}_i , \mathbf{U} \mathbf{U}^\top \bm{Y}_j + \sigma \bm{N}_j$ with $\bm{N}_i$ sampled i.i.d. $\sim \mathsf{N}(0, \sigma^2 \mathbf{I})$ and $\bm{N}_j$ sampled i.i.d. $\sim \mathsf{N}(0, \sigma^2 \mathbf{I})$. Denoting by $\tilde{\bm{X}_i}({\mathbf{U}}) = \mathbf{U} \mathbf{U}^\top \bm{X}_i + \sigma \bm{N}_i$, and by $\tilde{\bm{Y}}_j({\mathbf{U}}) = \mathbf{U} \mathbf{U}^\top \bm{Y}_j + \sigma \bm{N}_j$ the projected rank energy statistics is defined as: 

 \begin{align}\label{pre}
    \mathsf{PsRE}_{m,n}^2 \doteq   \max_{\mathbf{U} \in \mathbb{R}^{d \times k}: \mathbf{U}^\top \mathbf{U} = \mathbf{I}_k} & \left(\frac{2}{mn} \sum_{i=1}^{m} \sum_{j=1}^{n} \| \hat{\mathbf{R}}_{m,n}(\tilde{\bm{X}_i}({\mathbf{U}})) - \hat{\mathbf{R}}_{m,n}(\tilde{\bm{Y}}_j({\mathbf{U}})) \|  \right. \notag \\
    & \left.- \frac{1}{m^2} \sum_{i,j=1}^{m} \| \hat{\mathbf{R}}_{m,n}(\tilde{\bm{X}_i}({\mathbf{U}})) - \hat{\mathbf{R}}_{m,n}(\tilde{\bm{X}_j}({\mathbf{U}}))\| \right.\notag \\ & \left.- \frac{1}{n^2}  \sum_{i,j=1}^{n} \| \hat{\mathbf{R}}_{m,n}(\tilde{\bm{Y}}_i({\mathbf{U}})) - \hat{\mathbf{R}}_{m,n}(\tilde{\bm{Y}}_j({\mathbf{U}}))\|\right),
\end{align}
where the Ranks $\hat{\mathbf{R}}_{m,n}(\cdot)$ are computed using the samples $\tilde{\bm{X}_i}({\mathbf{U}}), \tilde{\bm{Y}}_j({\mathbf{U}})$ where $i = 1,2, \cdots, m, j= 1,2, \cdots, n$, and the Halton sequences in the original dimensions. The Projected soft Rank Energy, denoted $\mathsf{PsRE}_{m, n}^2$, is defined in an analogous manner as $\mathsf{sRE}$ using Equation \eqref{sre} and we leave it to the reader to observe it.

\section{Experimental Results}
\label{sec:sims}
In this section, we will compare the proposed scaled $\mathsf{sRE}^2_{m,n}$ statistics to the scaled $\mathsf{RE}^2_{m,n}$ based on the synthetic data. Going forward for the sake of brevity we refer to scaled $\mathsf{sRE}^2_{m,n}$ and scaled $\mathsf{RE}^2_{m,n}$ as $\mathsf{RE}$ and $\mathsf{sRE}$ respectively. We will illustrate the behavior of $\mathsf{sRE}$ statistics under the null and show experimentally that our proposed  statistics can achieve distribution-freeness with a small regularizer-based OT plan. We will demonstrate how $\mathsf{sRE}$ statistics depend on the optimal transport regularization parameter and find out which parameters achieve the notion of multivariate rank as closely as possible to RE \cite{deb2019multivariate}. We will also observe the effect of sample dimension on $\mathsf{sRE}$ statistics and show comparatively better results using projected soft rank energy \ref{proj_sec} for some specific distributional settings in higher dimensions. Finally, we will use $\mathsf{sRE}$ statistics in an application e.g. change point detection on some real data.

\noindent \textbf{Reproducible research}: All codes are available at  \url{https://github.com/ShoaibBinMasud/soft_projected_multivariate_rank.git}.

\subsection{Synthetic data}\label{synthetic}
We consider the following distributional settings to observe the behaviour of $\mathsf{sRE}$ statistics. We draw $m$ and $n$ i.i.d. samples of $d$ dimensional vectors $\bm X = (X_1, \dots, X_d)$ and $\bm Y = (Y_1, \dots, Y_d)$ respectively for each setting and carry-out two-sample goodness of fit test based on these observations. Throughout the experiments we assume $m = n$.

\begin{itemize}
    \item [($\text{v}1$)] $ (X_1, \dots, X_d) \overset{i.i.d.}{\sim} \; \text{cauchy}(0, 1)\; (Y_1, \dots, Y_d) \sim \text{Cauchy}(0.2, 1)$
    
    \item [($\text{v}2$)] $ X_1, Y_1 \overset{i.i.d.}{\sim} \mathcal U[0, 1]$, $X_k =  0.25 + 0.35 \times X_{k-1} + U_k, \; Y_k =  0.25 + 0.5 \times Y_{k-1} + U_k$ for $k = 2, \dots, d$, where $U_2, \dots, U_d \overset{i.i.d.}{\sim} \mathcal U[0, 1]$
    
    \item [($\text{v}3$)] $ \bm X \sim \mathcal N_d( 0, \Sigma_X)$ and $\bm Y \sim \mathcal N_d( 0, \Sigma_Y)$, where $\Sigma_X = 0.35 ^{|i - j|}$ and $ \Sigma_Y = 0.65 ^{|i - j|}$ for $1 \leq i, j \leq d $

     \item [($\text{v}4$)] $ \bm X \sim \mathcal N_d( 0, \Sigma_X)$ and $\bm Y \sim \mathcal N_d( 0, \Sigma_Y)$, where $\Sigma_X(i, i) = 1$ and $\Sigma_X(i, j) = 0.2$ for $i \neq j$ and $\Sigma_Y(i, i) = 1$ and $\Sigma_Y(i, j) = 0.5$ for $i \neq j$ for $1 \leq i, j \leq d$
    
     \item [($\text{v}5$)] $ \bm V \sim \mathcal N_d( 0, \Sigma_V)$ and $\bm W \sim \mathcal N_d( 0, \Sigma_W)$, where $\Sigma_V = 0.35 ^{|i - j|}$ and $ \Sigma_W = 0.65 ^{|i - j|}$ for $1 \leq i, j \leq d $. Set $X_i = \text{exp}(V_i)$ and $Y_i = \text{exp}(V_i)$ for $i =1, \dots, d$
  
     \item [($\text{v}6$)] $ \bm V \sim \mathcal N_d( 0, \Sigma_V)$ and $\bm W \sim \mathcal N_d( 0, \Sigma_W)$, where $\Sigma_V(i, i) = 1$ and $\Sigma_V(i, j) = 0.75$ for $i \neq j$ and $\Sigma_W(i, i) = 1$ and $\Sigma_W(i, j) = 0.5$ for $i \neq j$ for $1 \leq i, j \leq d$. Set  $X_i = \text{exp}(V_i)$ and $Y_i = \text{exp}(V_i)$ for $i =1, \dots, d.$

     \item [($\text{v}7$)] $ \bm X \sim \mathcal N_d(\bm \mu_X, 3\mathbf I)$ and $\bm Y \sim \mathcal N_d(\bm \mu_Y, 3\mathbf I)$ where $\bm \mu_X = (0, \dots, 0)$ and $\bm \mu_Y = (0.25, \dots, 0.25)$
   
     \item [($\text{v}8$)] $ X_1, \dots, X_d, V_1, \dots, V_d \overset{i.i.d.}{\sim} \text{Gamma}(2, 0.1)$ and $W_1, \dots, W_d = \text{exp}(Z)$ where $Z\sim \mathcal N(0, 1)$. Set $Y_i = V_i W_i$ for $i = 1,\dots, d$ and $A\sim \text{Ber}(0.8)$.
     
     \item [($\text{v}9$)]  $ \bm Z_1, \bm Z_2 \overset{i.i.d.}{\sim}\mathcal N_d(0, \mathbf I)$.  Let $\bm W = (W_1, W_2,\dots, W_d) \overset{i.i.d.}{\sim} \mathcal U(10)$. Finally, set $\bm X := \bm Z_1$ and $\bm Y := A\mathbf Z_2 + (1 - A) \bm W$
     
    \item [($\text{v}10$)] $ \bm Z_1, \bm Z_2 \overset{i.i.d.}{\sim}\mathcal N_d(0, \mathbf I)$.  Let $\bm W = (W_1, W_2,\dots, W_d) \overset{i.i.d.}{\sim} \mathcal N(10, 0.1)$. Set $\bm X := \bm Z_1$ and $\bm Y := A\bm Z_2 + (1 - A) \bm W$.
   
    \item [($\text{v}11$)] $\bm X \sim \text{Laplace}(\bm 0, \mathbf I_d) )$ and $\bm Y \sim \text{Laplace}(\bm 0, \mathbf I_d) )$ where $\bm \mu = (1, 0, \dots, 0)$
   
    \item [($\text{v}12$)] $X_1, \dots, X_d\sim \otimes_{i = 1}^{d-1} \mathcal N(0, 1) \otimes \mathcal N(0, 4)$ and $Y_1, \dots, Y_d \sim \otimes_{i = 1}^d \mathcal N(0, 1)$
\end{itemize}
\vspace{3mm}

Many of the distributional settings considered above are directly taken from \cite{deb2019multivariate,wang2020two}. For instance $\text{(v1)-(v10)}$ are adopted from \cite{deb2019multivariate}. $\text{(v11), (v12)}$ are slightly modified versions of similar settings in \cite{ramdas2015decreasing}. (v1) and (v8) are heavy-tailed Cauchy and Gamma distribution respectively with no finite first moment. In (v9) and (v10), $\bm X$ is a multivariate Gaussian, whereas $\bm Y$ is a multivariate Gaussian with a small percentage of Uniform and Gaussian noise respectively. (v3)-(v8) and (v12) are multivariate normals and log-normals. (v2) represents the auto-regressive model whereas in (v11) both $\bm X, \bm Y$ are sampled from the Laplace distribution.

\subsection{Distribution of \textrm{sRE} statistics under the null}\label{null_dist}

\begin{figure}[ht]
    \centering
    \includegraphics[width = \linewidth]{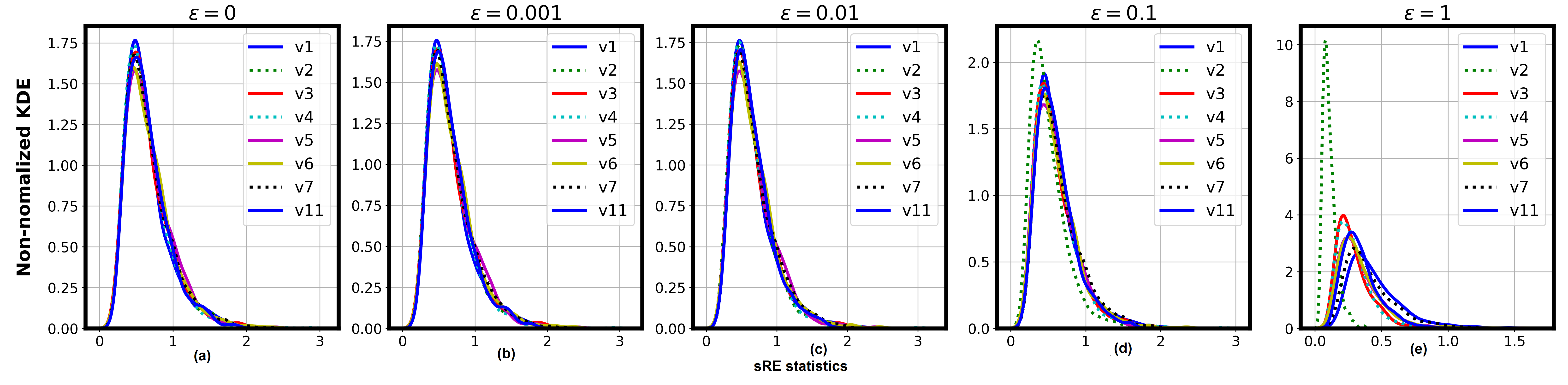}
    \vspace{-7mm}
    \caption{The estimated kernel density of soft rank energy with different regularizer $\epsilon = 0, 0.001, 0.01,0.1, 1$ under the null for $m=n=200$ and $d=3$.$y$-axis represents non-normalized kernel density estimated value $f(x)$.}
    \label{prof_null_dist}
\end{figure}

To illustrate the distribution of $\mathsf{sRE}$ statistics  under the null, we choose (v1)-(v7) and (v11) distributional settings, that include a heavy-tailed Cauchy, multivariate normal, log normal and Laplace distribution. We draw $200$ i.i.d. samples of $3$ dimensional vectors $\bm X =(X_1, X_2, X_3)\sim \mathsf P_X$ and $\bm Y =(Y_1, Y_2, Y_3)\sim \mathsf P_Y$, assuming $ \mathsf P_X \overset{d}{=} \mathsf P_Y$. For each setting, non-normalized kernel density of $\mathsf{sRE}$ statistics is estimated using $1000$ replicates of $\mathbf X, \mathbf Y\in \mathbb R^{200 \times 3}$, with different regularizers, $\epsilon = 0, 0.001, 0.01, 0.1, 1$ (Figure \ref{prof_null_dist}), where $\epsilon = 0$ refers to $\mathsf{RE}$ which guarantees exact distribution-freeness \cite{deb2019multivariate}.

Estimated $\mathsf{RE}$ ($\epsilon = 0$) statistics densities appear to be similar (Figure \ref{prof_null_dist}(a)) for (v1)-(v7) and (v11), that proves the distribution-freeness under the null as claimed in \cite{deb2019multivariate}. We observe the analogous density estimates of $\mathsf{sRE}$ statistics with $\epsilon = 0.001, 0.01$ (Figure \ref{prof_null_dist}(b)-(c)) for all settings, which also resemblance the density pattern obtained  using $\epsilon = 0$. As, $\epsilon$ increases e.g. $\epsilon = 0.1, 1$, estimated density curves no longer follow the same shape for different distributional settings (Figure \ref{prof_null_dist}(d)-(e)). Since entropic schemes for optimal transport with small regularizers converge to the Monge-Kantorovich problem \cite{carlier2017convergence}, transport plans using $\epsilon = 0.001, 0.01$ preserve the concept of multivariate rank \cite{deb2019multivariate}. However, with larger $\epsilon$, the transport plan gets more diffused, and $\mathsf{sRE}$ statistics digress largely from $\mathsf{RE}$ statistics.

\subsection{Effect of OT regularizer on sRE statistics}
\begin{figure}[ht]
    \centering
    \includegraphics[width = \linewidth]{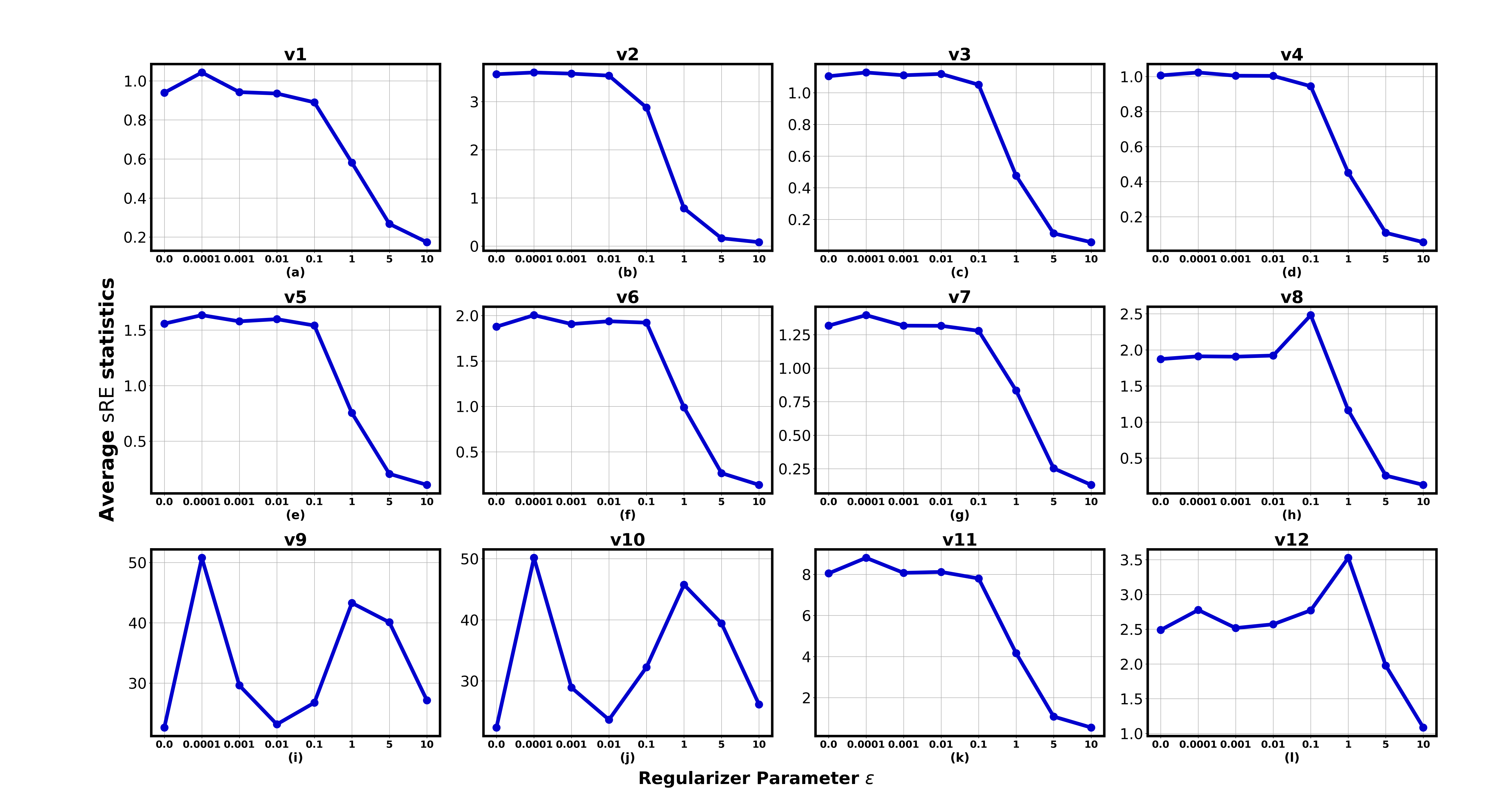}
    \vspace{-7mm}
  \caption{Average soft rank energy statistics on $y$-axis with different regularizers $\epsilon$ on $x$-axis computed over $500$ replicates for each distributional setting having $m= n= 200$ and $d = 3$.}
    \label{re_vs_sre}
\end{figure}

In this subsection, we observe how $\mathsf{sRE}$ statistics vary with respect to the optimal transport regularization parameter $\epsilon$ under alternative hypothesis. To analyze the relation,  $3$-dimensional, $m=n=200$ i.i.d. samples of $\bm X\sim \mathsf P_X$, $\bm Y \sim \mathsf P_Y$  are drawn for each setting listed in \ref{synthetic} assuming $\mathsf P_X\neq \mathsf P_Y$. We compute the average $\mathsf{sRE}$ statistics with different regularizers $\epsilon = 0.0001, 0.001, 0.01, 0.1, 1, 5, 1$, using $500$ replicates of $\mathbf X\in \mathbb R^{m\times d}, \mathbf Y\in \mathbb R^{n\times d}$ (see Figure \ref{re_vs_sre}), where $\epsilon = 0$ simply represents the $\mathsf{RE}$ statistics. Average $\mathsf{sRE}$ statistics for the distributional settings (v1)-(v8), (v11) and (v12) with $\epsilon = 0.0001, 0.001, 0.01$, are approximately equivalent to RE statistics (Figure \ref{re_vs_sre}(a-h,k,l)). However, when $\epsilon$ increases e.g. $\epsilon =0.1, 1, 5, 10$, average $\mathsf{sRE}$ decreases rapidly. Since the optimal transport plan gets diffused with  larger $\epsilon$, soft ranks obtained by multiplying the row-normalized plan with halton sequences (Equation \ref{softrank}) take approximately the same values for the samples from different distributions, thus weakening the notion of multivariate ranks that generally assign unique values to the order statistics. For (v9) and (v10), multivariate Gaussian distributions with small fraction of uniform and Gaussian noise respectively, average $\mathsf{sRE}$ fluctuates with $\epsilon$ (Figure \ref{re_vs_sre}(k)-(l)). However, even with large regularizers $\epsilon = 0.1, 1, 5, 10$, average $\mathsf{sRE}$ is still comparable to the $\mathsf{RE}$ statistics. To be noted here, both $\mathsf{RE}$ and $\mathsf{sRE}$ are not robust to noise. Additionally, a minimal shift either in mean or covariance in one of the dimensions e.g. (v11), (v12) may yield higher  $\mathsf{RE}$ and $\mathsf{sRE}$ statistics, thus rejecting the true null hypothesis and raising false alarms in many practical applications e.g. change point detection. Future research can be carried out to make hypothesis tests based on $\mathsf{sRE}$ statistics more robust to noise and outliers. 

\subsection{sRE   with respect to different sample dimensions}\label{dim_sRE}

\begin{figure}[ht]
    \centering
    \includegraphics[width = \linewidth]{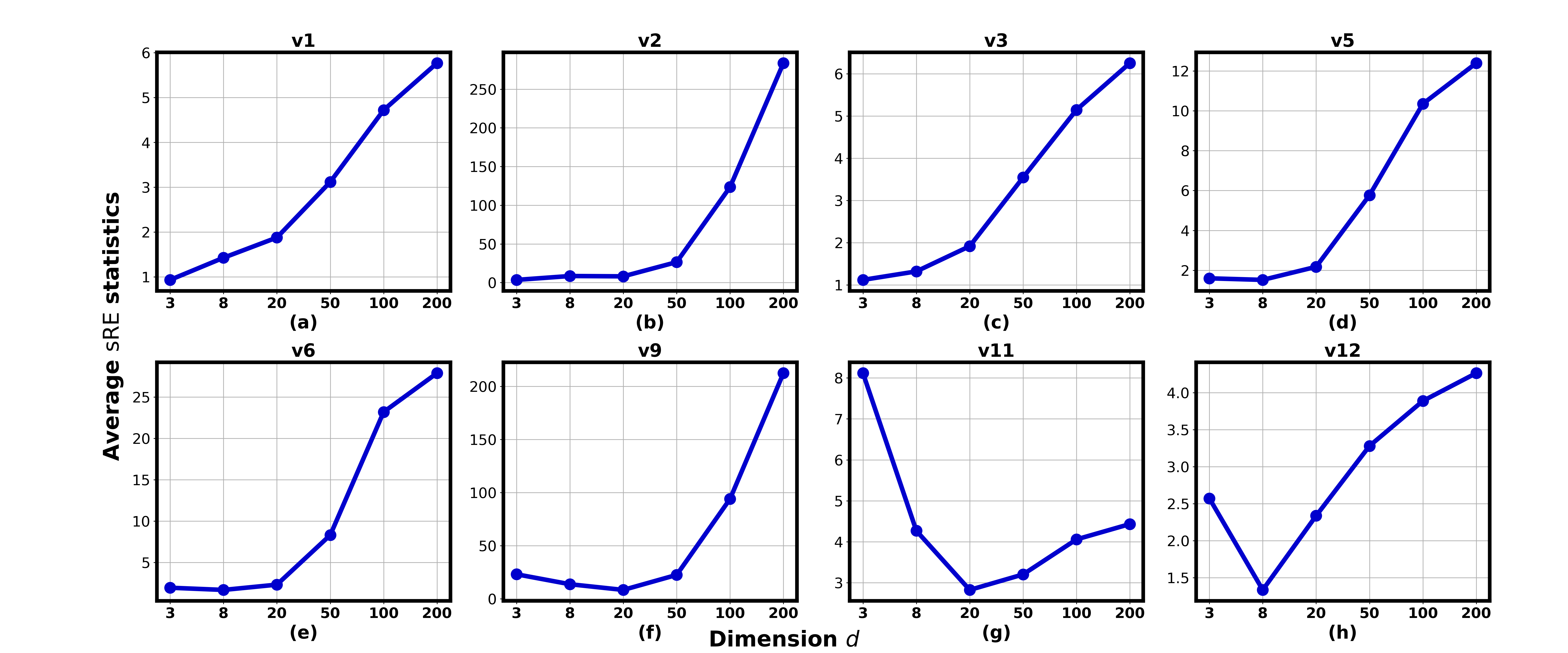}
    \vspace{-7mm}
    \caption{Average soft rank energy statistics on $y$-axis with respect to different dimensions $d\in D = \{3, 8, 20, 50, 100, 200\}$, with  sample size $m = n= 200$.}
    \label{dim}
\end{figure}

We investigate the concept of multivariate soft rank in higher dimension  and assess the relationship between $\mathsf{sRE}$ statistics and the dimension, for $d>2$.
We draw $m=n=200$ i.i.d. samples  of $\bm X\in P_X$ and  $\bm Y\sim P_Y$, $P_X\neq P_Y$, of dimension $d\in D$, where $D =\{3,8, 20, 50, 100, 200$\}, for distributional settings (v1)-(v3), (v5), (v6), (v9), (v11) and (v12). For every combination of $m$ and $d$, average $\mathsf{sRE}$ statistics are computed using $500$ replicates of $\mathbf {X, Y}\in \mathbb R^{m\times d}$  with regularizer $\epsilon = 0.01$  (Figure \ref{dim}). $\epsilon= 0.01$ is chosen since $\mathsf{sRE}$ statistics imitate the distribution-free behaviour under the null, providing values comparable to $\mathsf{RE}$ (Figures \ref{prof_null_dist}, \ref{re_vs_sre}) with less computational complexity. Average $\mathsf{sRE}$ statistics show positive correlation with the dimension for (v1), (v3), (v5) and (v6) (Figure \ref{dim}(a, c, d, e)). We observe a similar trend for the autoregressive setting (v2) (Figure \ref{dim}(b)). However, for (v9) which is a multivariate Gaussian contaminated with uniform noise, $\mathsf{sRE}$ statistics  first decreases slightly and then increases at a much higher rate when $d>20$ (Figure \ref{dim}(f)). For (v11), we observe a sharp decrease in average $\mathsf{sRE}$ but it starts increasing after $d >20$, though $\mathsf{sRE}$ statistics is still small compared to values in smaller dimension. We notice  a phenomenon for (v12) alike to (v11) with an elbow at $d = 8$ (Figure \ref{dim}(h)), yet small compared to other normal and lognormal settings e.g. (v3), (v5) and (v6) in higher dimensions. A small shift in mean and variance in one particular dimension, while preserving the same mean and variance along all the other dimensions of $\bm{Y}$ in (v11) and of $\bm X$ in (v12), explains such behaviour of average $\mathsf{sRE}$ statistics. Therefore, in high dimension, $\mathsf{sRE}$ based hypothesis test might consequently results into frequent rejection of true null hypothesis for similar structures e.g. (v11) and (v12).

\subsection{Projected Soft Rank Energy}
\begin{figure}[ht]
    \centering
    \includegraphics[width = 16cm, height = 6cm]{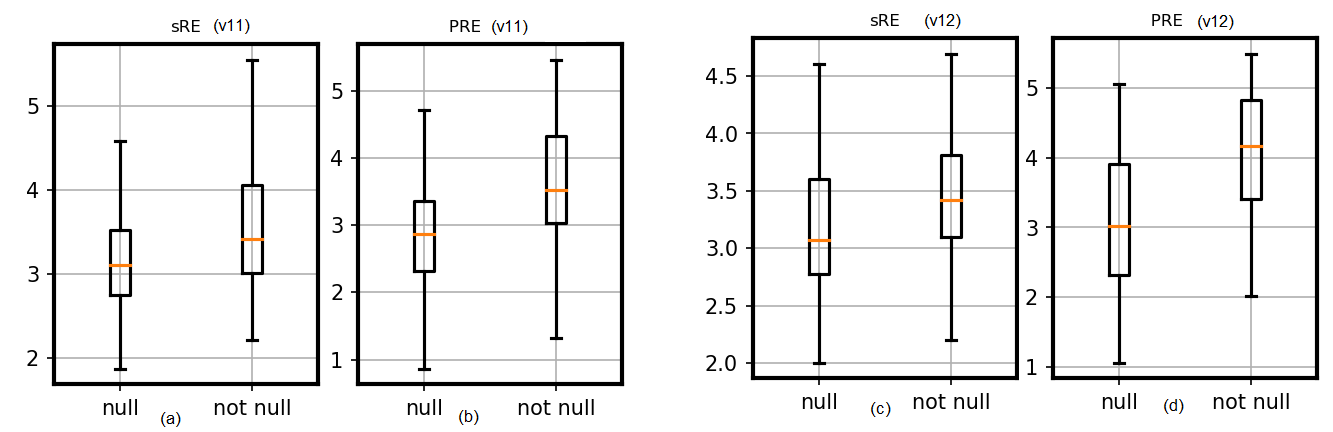}
    \caption{Box and whisker plot of $\mathsf{sRE, \text{scaled}\;PsRE}$ under both null and alternative hypotheses for two different sets of distribution with $\epsilon = 0.001$.}
    \label{proj8}
\end{figure}

We compare the scaled projected soft rank energy $\frac{k}{d}\cdot\mathsf{PsRE}^2$ (henceforth referred to as $\mathsf{PsRE}$ for brevity) proposed in Section \ref{proj_sec} where $d$ and $k$ represents the dimension of the original sample space and the subspace the samples projected onto respectively to $\mathsf{sRE}$ for the settings (v11) and (v12). In higher dimensions, for these cases, $\mathsf{sRE}$ based hypothesis tests might not be as useful as in lower dimensions since $\mathsf{sRE}$ statistics do not exhibit the typical positive dependency with the dimension (see Figure \ref{dim}). To check whether $\mathsf{PsRE}$ provides more distinguishable statistics between the null and alternative hypothesis than $\mathsf{sRE}$, we draw $d = 100$-dimensional $m=n=200$ i.i.d. samples of $\bm X\in \mathsf P_X$ and $\bm Y\sim \mathsf P_Y$ when $\mathsf P_X = \mathsf P_Y$ (null hypothesis) and $\mathsf P_X \neq \mathsf P_Y$ (alternative hypothesis). First, we compute the average $\mathsf{sRE}$ statistics in $d = 100$ and then project $\mathbf{X, Y}\in \mathbb R^{m\times 100}$ to $k = 3$-dimensional subspace, $\mathbf{X_{\text{proj}}, Y_{\text{proj}}}\in \mathbb R^{200\times 3}$, using a Python manifold optimization  package called as \emph{Pymanopt} and compute the $\mathsf{sRE}$ statistics  with regularizer $\epsilon = 0.001$, under both null and alternative hypothesis (Figure \ref{proj8}). We observe that under the null and alternative hypotheses, average $\mathsf{sRE}$ statistics are very similar when $d=100$ both for (v11) and (v12) (Figure \ref{proj8}(a, c)), which consequently leads to poor power. On the contrary, average scaled $\mathsf{PsRE}$ statistics has a more distinguishable median for both (v11), (v12) (Figure \ref{proj8}(b ,d))  which eventually ensures greater power to GoF test with a compromise to produce few false alarms.

\subsection{Algorithm for Change Point Detection based on proposed sRE}
In this section, we incorporate the idea of $\mathsf{sRE}$ statistics in a widely studied two-sample test e.g. change point detection \cite{aminikhanghahi2017survey}, and evaluate the performance of $\mathsf{sRE}$ statistics on a real multivariate  time-series dataset. Following the recent works in \cite{cheng2020optimal, OJSP2020}, we apply $\mathsf{sRE}$ as a test statistic to detect the change points on the dataset utilizing a sliding window based two-sample test. Given a time series data $Z[t] \in \mathbb R^d$, $t = 1, \dots, T$,  consists of distinct segments $[0, \tau_1], \; [\tau_1 +1, \tau_2], \dots, [\tau_k + 1, T]$ with $\tau_1 < \tau_2 < \dots$, coming from unknown  distributions where $\tau_1, \tau_2, \dots, \tau_k$  are referred as the change points. Unsupervised change point detection problem aims to estimate $\tau_1, \tau_2, \dots, \tau_k$ without any prior knowledge of the underlying distribution of distinct time segments and the number change points. Given a window size $n$, we define two different time segments at time $t$ by taking samples before $t$ as $\mathbf X = \{Z[t -n], Z[t -n +1], \dots, Z[t -1]\} \in \mathbb R^{n\times d}$ and after $t$ as $\mathbf Y = \{Z[t + n], Z[t + n  - 1], \dots, Z[t +1]\}\in \mathbb R^{n\times d}$. We assume both $\mathbf {X, Y}$ are sampled from unknown distributions $\bm{\mu, \nu}$ respectively. Similar to hypothesis testing set up, we reject the null hypothesis $H_0$ if $\bm \mu \neq \bm\nu$ and will define time $t$ as a change point.

\subsubsection{Performance on real data}
\begin{figure}[ht]
     \centering
  \includegraphics[width = \linewidth]{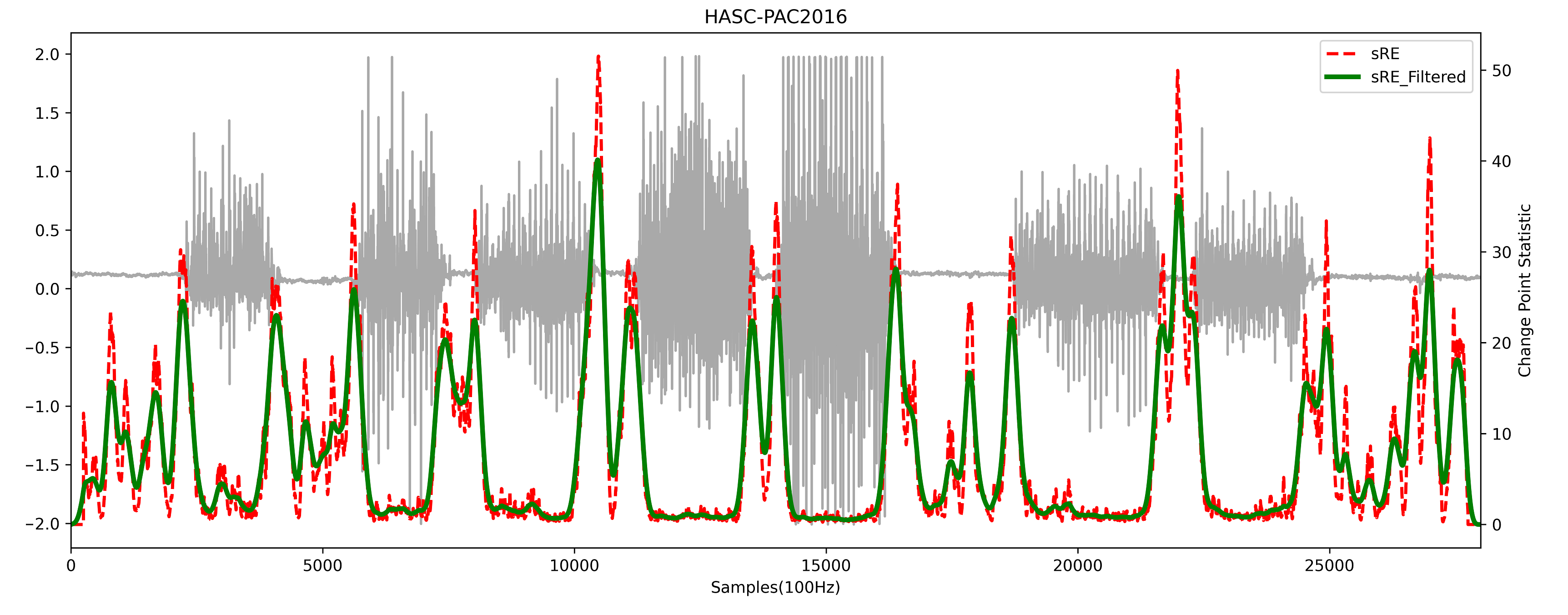}
     \caption{Sample output for HASC-PAC2016 human activity accelerometer data sequence (grey) with the filtered (solid green) and unfiltered (dashed red) sRE statistics. It appears that  our method results in frequent false alarms in support of a change point whereas HSAC does not consider that as a  true change point.}
     \label{fig:cpd}
\end{figure}

We applied our proposed computationally non-parametric $\mathsf{sRE}$ statistics based change point detection on \textbf{HASC-PAC2016} dataset, which consists of over $700$ three-axis accelerometer sequences of subjects performing six types of actions: stay, walk, jog, skip, stairs up and stairs down. We only have taken one longest sequence where all of the actions were present. We select the window size $n =250$ and compute the $\mathsf{sRE}$ statistics with $\epsilon = 0.01$. We then convolve the $\mathsf{sRE}$ statistics  with a low-pass filter as proposed in \cite{cheng2020optimal} (Figure \ref{fig:cpd}).

Our method is able to detect all the change points that are considered as real change points in the dataset using both filtered and unfiltered $\mathsf{sRE}$ statistics. However, this method provides frequent false alarms. This is because $\mathsf{sRE}$ is very sensitive and can detect if there exists any small shift either in mean or covariance between the adjacent windows. The analysis of these properties and mitigating false alarms in presence of anomalies is an important direction of future work. 
 
\section*{Conclusion and Future Work}
We have extended the concept of optimal transport-based multivariate rank to soft multivariate rank utilizing entropic regularization.  Experimental results show that soft rank energy is approximately distribution-free under the null and behaves similarly to original OT-based multivariate rank energy \cite{deb2019multivariate} for small regularization parameters. The proposed soft-rank energy allows us to consider a projected version of the rank-energy tests for high-dimensional data. For several distributional settings, our proposed projected soft rank energy performs better than rank energy and soft rank energy. Future research includes analyzing the theoretical properties of the proposed tests and addressing the sensitivity of these tests to the presence of outliers in the data.

\section*{Acknowledgment}
We acknowledge helpful technical discussions with Ahmed Abbasi and Ruijie Jiang. 


\bibliographystyle{./bibliography/IEEEtran}
\bibliography{IEEEexample.bib}

\end{document}